\documentclass[journal]{IEEEtran}
\ifCLASSINFOpdf
\else
\fi
\usepackage[utf8]{inputenc}
\usepackage{colortbl}
\usepackage[dvipsnames]{xcolor}
\usepackage{epsfig}
\usepackage{graphicx}
\usepackage{amsmath}
\usepackage{amssymb}
\usepackage{algorithm}
\usepackage{subfigure}
\usepackage{textcomp}
\usepackage{placeins}
\usepackage{epstopdf}
\usepackage{siunitx}
\usepackage{makecell,multirow}
\usepackage{arydshln}
\usepackage{adjustbox}
\usepackage{cite}
\usepackage{cuted}
\usepackage{scalerel,stackengine}
\stackMath
\newcommand\reallywidehat[1]{%
\savestack{\tmpbox}{\stretchto{%
  \scaleto{%
    \scalerel*[\widthof{\ensuremath{#1}}]{\kern-.4pt\bigwedge\kern-.4pt}%
    {\rule[-\textheight/2]{1ex}{\textheight}}
  }{\textheight}%
}{0.6ex}}%
\stackon[1pt]{#1}{\tmpbox}%
}
\usepackage{algpseudocode} 
\usepackage[pagebackref=true,breaklinks=true,colorlinks,bookmarks=false]{hyperref}
\begin{document}
%
\title{Bidirectional Mapping Coupled GAN for Generalized Zero-Shot Learning}
%
%
%
\author{Tasfia~Shermin,   Shyh Wei~Teng, Ferdous~Sohel,~\IEEEmembership{Senior Member,~IEEE,}   Manzur~Murshed,~\IEEEmembership{Senior Member,~IEEE,}   and~Guojun~Lu,~\IEEEmembership{Senior Member,~IEEE} 
	\thanks{Tasfia Shermin, Shyh Wei Teng, Manzur Murshed,and Guojun Lu are with the School of Engineering, Information Technology and Physical Sciences, Federation University, Australia (email:t.shermin@federation.edu.au)}
	\thanks{Ferdous Sohel is with Information Technology, Murdoch University, WA-6150, Australia}
\thanks{Under Review*}}
\markboth{}%
{Shell \MakeLowercase{\textit{et al.}}: Bare Demo of IEEEtran.cls for IEEE Journals}
%
\maketitle
\begin{abstract}
Bidirectional mapping-based generalized zero-shot learning (GZSL) methods rely on the quality of synthesized features to recognize seen and unseen data. Therefore, learning a joint distribution of seen-unseen domains and preserving domain distinction is crucial for these methods. However, existing methods only learn the underlying distribution of seen data, although unseen class semantics are available in the GZSL problem setting. Most methods neglect retaining domain distinction and use the learned distribution to recognize seen and unseen data. Consequently, they do not perform well. In this work, we utilize the available unseen class semantics alongside seen class semantics and learn joint distribution through a strong visual-semantic coupling. We propose a bidirectional mapping coupled generative adversarial network (BMCoGAN) by extending the coupled generative adversarial network into a dual-domain learning bidirectional mapping model. We further integrate a Wasserstein generative adversarial optimization to supervise the joint distribution learning. We design a loss optimization for retaining domain distinctive information in the synthesized features and reducing bias towards seen classes, which pushes synthesized seen features towards real seen features and pulls synthesized unseen features away from real seen features. We evaluate BMCoGAN on benchmark datasets and demonstrate its superior performance against contemporary methods.
\end{abstract}

%
\IEEEpeerreviewmaketitle
\section{Introduction}	
Deep learning classifiers generally follow the setting of supervised classification, which means they expect the training and testing data from the same categories with a similar underlying distribution. This makes the task of recognizing images from categories that are unseen during training challenging for such models. Deep learning models also require a large number of labeled training data to achieve a satisfactory performance \cite{tsherminDAMC}. Although sufficient images of ordinary object categories are available, there are many object categories with limited or no visual data, such as endangered plants and animals \cite{ding2019selective, elhoseiny2017link,zhao2019recognizing, zheng2017learning, lampert2013attribute, xian2017zero, zhu2018generative}. To solve these problems, \textit{Zero-shot learning} (ZSL) methods are studied. In particular, ZSL aims to learn visual classifiers for the unseen categories that have no available or labeled visual data by relating the semantic descriptor space of the seen and unseen categories \cite{han2020learning}. This implies that the ZSL setting supports the availability or generation of the semantic attributes of the unseen categories. For example, birds have some common attributes such as `beak shape', `bill shape', `belly color', `head shape', and `tail color'. The semantic attributes of endangered birds can easily be generated following the common attribute fields.

ZSL methods have two main categories: inductive (no images of unseen classes are used during training) \cite{lampert2009learning, lampert2013attribute, farhadi2009describing} and transductive (unlabelled images of unseen classes are used during training) \cite{fu2015transductive, yu2018transductive, jiang2019transferable}. ZSL methods aim to test the trained visual classifier on images from unseen classes only. The ZSL setting relies on the assumption that the trained classifier knows whether a test image comes from the seen or unseen domain. This assumption is unrealistic. Therefore, the ZSL setting is extended to a more realistic setting called Generalized Zero-Shot Learning (GZSL) \cite{ni2019dual, jiang2019transferable, min2020domain, geng2020guided, han2020learning}, where the classifier needs to classify images from both seen and unseen classes during testing. In this paper, we follow the inductive GZSL setting i.e, we do not utilize images from unseen classes during training. 

Latent space embedding learning and feature synthesizing are the two most studied categories of ZSL methods. Embedding learning methods explore the visual-semantic relationship of seen classes to extend the traditional classification task to unseen classes with no available training images \cite{farhadi2009describing, lampert2009learning, lampert2013attribute, akata2015evaluation, frome2013devise, yu2018zero, socher2013zero, liu2018generalized, zhang2019co, xie2019attentive}. Most existing methods either maps the visual features to semantic space \cite{akata2013label, huang2019generative, xian2016latent, romera2015embarrassingly, frome2013devise, akata2015evaluation, kodirov2017semantic} or semantic attributes to visual space \cite{shigeto2015ridge, dinu2014improving, zhang2017learning}. Other methods focus on attention-based learning \cite{huynh2020fine, ji2018stacked, huynh2020sha, zhu2019semantic, shermin2020integrated}. Mapping high dimensional visual features to low dimensional semantic space may reduce the variance of features and create hubness problem \cite{shigeto2015ridge, dinu2014improving, zhang2017learning}. Similarly, semantic to visual mapping is not optimal as one class may have several corresponding visual features \cite{huang2019generative}.

On the other hand, feature synthesizing or generative methods only rely on the class semantics for adversarially generating visual features \cite{kumar2018generalized, han2020learning, felix2018multi, schonfeld2019generalized, xian2018feature, li2019leveraging}. However, the generative methods capture the visual distribution only via a unidirectional alignment from the class semantics to the visual feature. This leads to weak visual-semantic coupling, which is vital for zero-shot tasks\cite{huang2019generative, ni2019dual, chandhok2020enhancing}, and harms the performance. To handle this issue and improve performance, recent methods \cite{ni2019dual, huang2019generative, xing2020robust, pambala2020generative, liu2020information, gu2020generalized} use the bidirectional mapping between visual and semantic domains.

For GZSL, during testing, the bidirectional mapping models have to recognize both seen and unseen classes. Thus, integrating the knowledge of the joint distribution of seen-unseen domains will enhance the GZSL recognition performance. However, existing models ignore learning seen-unseen domains joint distribution \cite{ni2019dual, huang2019generative, xing2020robust, pambala2020generative, liu2020information, gu2020generalized}. In addition, preserving inter-domain discriminative information in the model is necessary for improved classification and limiting bias towards seen classes during testing. However, except for a few existing bidirectional mapping methods \cite{pambala2020generative, liu2020information}, they \cite{ni2019dual, chandhok2020enhancing, huang2019generative, xing2020robust, gu2020generalized} ignore the challenge of reducing bias towards seen classes while performing zero-shot recognition (for more details see section~\ref{bi}). 

In this work, we propose a new generative bidirectional mapping GZSL method to address the above-mentioned issues. Our ultimate aim is to learn two broad tasks: 1) learn the joint distribution of seen-unseen domain by establishing strong coupling between the semantic-visual spaces of both domains through bidirectional mapping and 2) preserve domain discriminative characteristics in the generated feature space. Task 1 will enhance seen-unseen data classification performance, and Task 2 will mitigate bias towards seen classes.

As we follow the inductive setting, we do not have access to unseen image data. However, the GZSL problem setting states the availability of unseen class semantics \cite{ni2019dual, han2020learning}. Therefore, we aim to learn the joint distribution using seen and unseen class semantics. The Coupled GAN (CoGAN) \cite{liu2016coupled} has a generative and weight sharing structure. So, it can learn dual-domain joint distribution with just samples drawn from the marginal distributions by restricting the network capacity. Therefore, to learn Task 1, we propose to extend the Coupled GAN (CoGAN) \cite{liu2016coupled} into a seen-unseen domain joint distribution learning bidirectional mapping method. We refer to our proposed method as Bidirectional Mapping Coupled Generative Adversarial Network (BMCoGAN). We partially adopt the weight sharing properties of CoGAN in BMCoGAN to learn the joint distribution and integrate the bidirectional mapping property in the CoGAN structure by introducing regressors (Fig.~\ref{f1}). To the best of our knowledge, ours is the first attempt to utilize and extend CoGAN structure for GZSL recognition.

In principle, for both domains, BMCoGAN generates visual features from class semantics, reconstructs class semantics back from generated visual features, and adversarially assesses reconstructed class semantics against the real class semantics. Since the real images are not available for unseen classes, we hope to learn unseen feature generation and contribute to joint-distribution learning through this bidirectional mapping and adversarial semantic assessment. However, for seen classes, besides bidirectional mapping and semantic assessment, visual feature generation is adversarially supervised by available training images. 

To preserve discrimination among seen and unseen classes (Task 2), we propose to push generated seen visual features towards real seen visual features and pull generated unseen visual features away from real seen visual features. This mitigates the bias towards seen classes. 

The main contributions are as follows: \\
\textbf{1)} We propose a generative adversarial GZSL network, which learns seen-unseen domain joint-distribution through bidirectional mapping and generates better domain discriminative features.\\
\textbf{2)} To capture the dual-domain joint distribution, we propose to extend the CoGAN into a bidirectional mapping network (BMCoGAN) for GZSL tasks. In particular, we relate the visual-semantic spaces by imposing bidirectional mapping in CoGAN and modify the weight sharing properties of CoGAN. 
We also support the joint distribution learning with a supervisory Wasserstein generative adversarial optimization.\\ 
\textbf{3)} Unlike existing methods, in addition to distance-based optimization for bidirectional mapping, the proposed BMCoGAN adversarially ensures that the underlying distribution of the synthesized class semantics is aligned with the real semantics for both seen and unseen domains.\\
\textbf{4)} To reduce bias towards seen classes, we encourage the proposed method to maintain more similarity between real seen and synthesized seen features than real seen and synthesized unseen features. This helps in retaining the required discrimination between both domains.\\
\textbf{5)} We present an extensive empirical evaluation of BMCoGAN on several datasets to demonstrate its superior performance compared to contemporary GZSL methods.

The proposed BMCoGAN can be applied to many image processing applications such as image or video classification or recognition. Section II presents a brief discussion about contemporary GZSL methods. Details on BMCoGAN are described in Section III. Thorough performance analysis of BMCoGAN and comparison with contemporary approaches on various datasets are provided in Section IV.
\section{Related Work}
\label{related}
In this section, we provide a brief description of existing GZSL methods. Works on GZSL can broadly be divided into two categories: embedding learning methods and generative methods.
\subsection{Embedding Learning Methods} 
The embedding learning methods learn to project either visual features to the semantic space \cite{ huang2019generative, romera2015embarrassingly, akata2015evaluation, kodirov2017semantic} or semantics to the visual space \cite{shigeto2015ridge, dinu2014improving, zhang2017learning} based on seen classes for the GZSL task. Majority of the existing embedding learning GZSL methods use global visual features for GZSL task \cite{frome2013devise, akata2015label, hubert2017learning, romera2015embarrassingly, changpinyo2016synthesized, xian2016latent, zhang2018zero, liu2018generalized, zhang2017learning, akata2015evaluation}. To avoid noise and non-discriminative information in the embedding \cite{zhu2018generative, elhoseiny2017link, xian2016latent, zhang2016zero, changpinyo2016synthesized}, a few works have applied attention mechanisms \cite{akata2016multi, ji2018stacked, xie2019attentive, zhu2019semantic, guo2018zero, liu2019attribute, huang2019attention, huynh2020fine, shermin2020integrated}.

\subsection{Generative Methods}
Generative methods can be grouped into two categories as follows:
\subsubsection{Unidirectional Mapping Methods}
These methods adversarially learns to synthesize visual features from class semantics and reduce the GZSL to a standard supervised classification task \cite{xian2018feature, sariyildiz2019gradient, schonfeld2019generalized, huang2019generative, ni2019dual, kumar2018generalized, keshari2020generalized, yu2018zero, kumar2018generalized, xian2019f, felix2018multi}. For generation of unseen class features, f-clsWGAN \cite{xian2018feature}, CVAE \cite{felix2018multi, mishra2018generative}, SE-GZSL \cite{kumar2018generalized} used conditional Generative Adversarial Networks (GANs) or Variational Autoencoders (VAE).

However, the above-mentioned methods rely on only one-directional semantic to visual features mapping, which does not guarantee strong visual-semantic interactions. Therefore, bidirectional mapping methods are studied.

\subsubsection{Bidirectional Mapping Methods}
\label{bi}
Bidirectional mapping methods perform semantic-to-visual and visual-to-semantic mapping to ensure strong interaction between the visual and semantic spaces. This bond is vital for GZSL tasks. The proposed method falls under this category. 

The first effort in this field is DASCN \cite{ni2019dual}, which simply employes two generative adversarial networks to construct visual features from semantics and reconstruct back semantics from the visual space for dual learning. GDAN \cite{huang2019generative} utilizes a regressor instead of another GAN to map generated features back to the semantic space. For bidirectional optimization, GDAN minimizes the distance between the real and reconstructed semantics. This optimization is not strong enough to preserve high-level semantic information \cite{ni2019dual}. Unlike GDAN, in the proposed BMCoGAN, in addition to distance-based optimization, we supervise the semantic reconstruction through adversarial optimization using the coupled discriminators. GZSL-AVSI \cite{chandhok2020enhancing} proposes to implement dual learning between an inference module and a generative module. GZSL-AVSI uses a Wasserstein semantic alignment loss to optimize the semantic space. On the other hand, our proposed BMCoGAN optimizes Wasserstein distance-based adversarial loss for adversarially supervising the seen domain's visual feature space.

In addition to dual learning, RBGN \cite{xing2020robust} incorporates adversarial attack strategies to train a more attentive discriminator. VAEs rely only on the lower bound of the log-likelihood of observed data. To address this issue, cFLOW-ZSL \cite{gu2020generalized} utilizes generative flow to estimate accurate likelihood during the bidirectional mapping using VAEs. 

IBZSL \cite{liu2020information} uses the information bottleneck constraint and data uncertainty estimation technique to design a bias passing mechanism to alleviate the noises and gap between visual features and human-annotated semantics and hope to reduce seen domain bias. ISE-GAN \cite{pambala2020generative} proposed an integrated-classifier to be trained with the bidirectional learning scheme. The integrated-classifier is trained to reduce bias towards seen classes, and without the integrated-classifier, the SE-GAN \cite{pambala2020generative} is inclined towards the seen classes. On the contrary, we infuse the knowledge of smoothing out bias towards seen classes in the generator and the discriminator by designing a loss optimization function. Compared to IBZSL \cite{liu2020information}, the proposed BMCoGAN does not require expensive constraint computation to limit bias towards seen classes. And, unlike SE-GAN \cite{pambala2020generative}, our generator can preserve seen-unseen domain discrimination in the visual features.

\section{Proposed Method}
In this section, we formally describe the generalized zero-shot learning setting and present our proposed method.

\subsection{Problem Setting}
The GZSL setting has a seen $\mathcal{Y}^{s}$ domain and an unseen $\mathcal{Y}^{u}$ domain with $C^s$ and $C^u$ classes, respectively, where  $\mathcal{Y}^{s} \cap \mathcal{Y}^{u}=\emptyset$. $\{1, \ldots, C^s\}$ denotes the seen classes and  $\{C^s+1, \ldots, C^s+C^u\}$ denotes the unseen classes. The seen domain has $N$ labeled images, $\mathcal{D}^s=$ $\left\{\left(x_{i}, y_{i}\right) \mid x_{i} \in \mathcal{X}^s, y_{i} \in \mathcal{Y}^{s}\right\}_{i=1}^{N}$, $\mathcal{X}$ denotes the visual feature space. The unseen classes have no available training images. The seen \textit{class semantics} for $c \in \mathcal{Y}^{s}$ are $\mathcal{A}^s =\left\{a_{c}\right\}_{c=1}^{C^s}$. The unseen \textit{class semantics} for $c \in \mathcal{Y}^{u}$ are $\mathcal{A}^u =\left\{a_{c}\right\}_{c=C^s+1}^{C^u}$. The class semantics of class $c$ is ${a_c}=[a_c^1, \dots, a_c^A]$, where $a_{c}^A$ denotes the measure of the presence of the $A^{th}$ semantic attribute in the class. 
The goal of GZSL is to train visual classifiers for all seen and unseen classes $h_{gzsl}:\mathcal{X} \rightarrow \mathcal{Y}^{s} \cup \mathcal{Y}^{u}$. 

\subsection{Coupled Generative Adversarial Networks}
\begin{figure*}[!ht]
\centering
  \includegraphics[width=.7\textwidth,height=4cm]{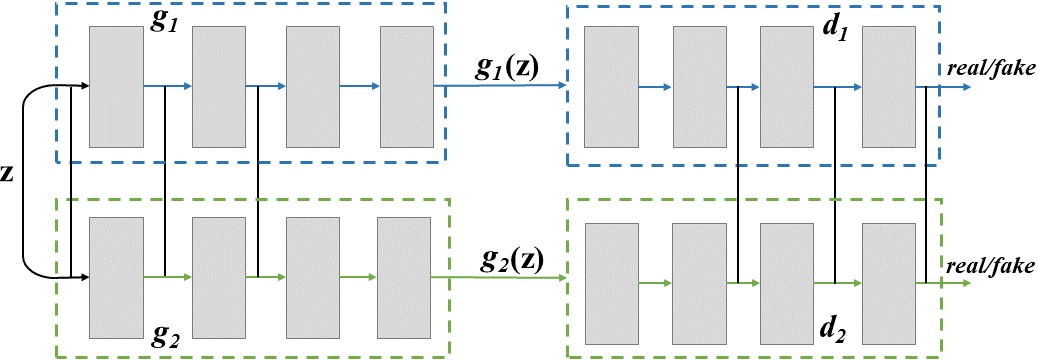}
\caption{Block diagram of CoGAN.}
\label{fcogan} 
\end{figure*}

In Coupled Generative Adversarial Networks (CoGAN) \cite{liu2016coupled}, there are two GANs with shared layers to handle high-level semantic features and separate layers to deal with low-level features for different domains. CoGAN also has two discriminators with shared layers, as shown in Fig.~\ref{fcogan}.
This setting allows CoGAN to capture the joint distribution and generate images of multiple domains. CoGAN can learn the joint distribution without relying on the correspondence between data samples in the two domains. The minimax objective function for CoGAN is as follows,
\begin{equation} \label{e1}
\begin{split}
\max_{g_{1}, g_{2}}\min_{d_{1}, d_{2}}\mathcal{L}&\equiv\mathbb{E}_{x_{1}\sim p_{x_{1}}}\left[\log d_{1}\left(x_{1}\right)\right]\\
&+\mathbb{E}_{z \sim p_{z}}\left[\log\left(1-d_{1}\left(g_{1}(z)\right)\right)\right]
\\&+\mathbb{E}_{x_{2} \sim p_{x_{2}}}\left[\log d_{2}\left(x_{2}\right)\right]\\
&+\mathbb{E}_{z \sim p_{z}}\left[\log \left(1-d_{2}\left(g_{2}(z)\right)\right)\right].
\end{split}
\end{equation}
Here, $\text{GAN}_i$ consists of generator $g_i$ and discriminator $d_i$, and $i= 1, 2$. $x_1$ and $x_2$ denote two samples from different domains, and $z$ represents the Gaussian noise distribution. The CoGAN optimization is subjected to two constraints as follows,

$\begin{array}{ll}\theta_{g_{1}^{j}}=\theta_{g_{2}^{j}} & 1 \leq j \leq s_{g} \\ \theta_{d_{1}^{l_{1}-k}}=\theta_{d_{2}^{l_{2}-k}} & 0 \leq k \leq s_{d}-1\end{array}$

where $\theta_{g_{i}^{j}}$ denotes the parameter of the $j^{th}$ layer from the top of the generator $g_i$, $\theta_{d_{i}^{l_{i}-k}}$ represents the parameter of the $(k+1)^{th}$ layer from the final layer of the discriminator $d_i$, $l_i$ denotes the number of layers in $d_i$. 
\begin{figure*}[!ht]
        \centering
        \includegraphics[width=1\textwidth,height=5cm]{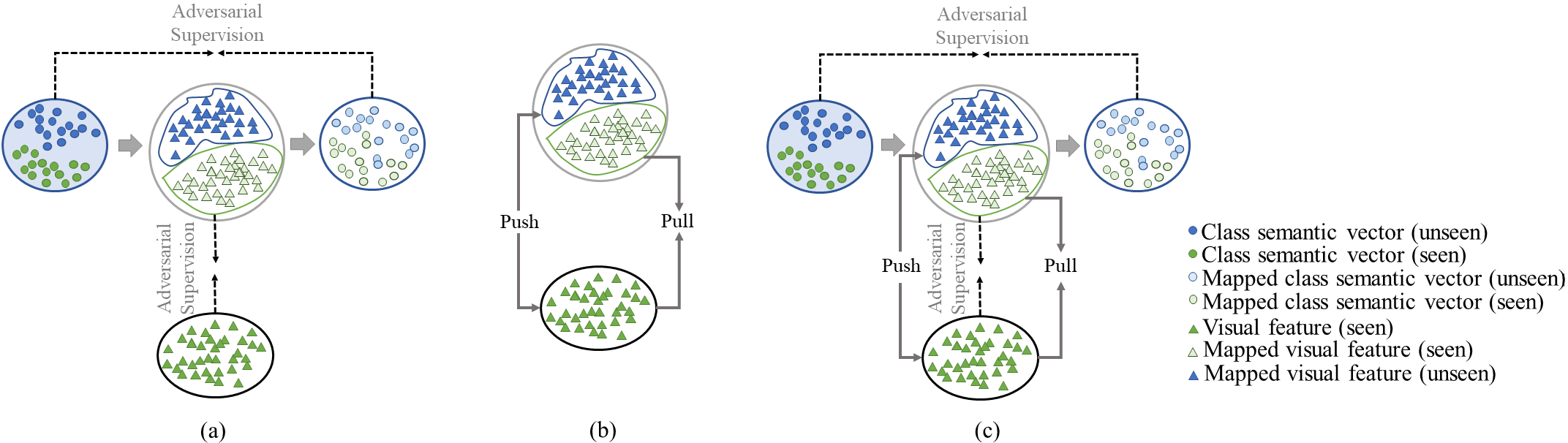}
        \caption{A conceptual overview of the proposed method. (a) To learn a joint distribution of seen and unseen domains, we learn to construct visual features from class semantics and reconstruct class semantics back from the generated visual features. This bidirectional mapping ensures a strong relation between semantic-visual space. This compensates for the absence of supervision from unseen real features during synthetic unseen feature generation. 
        (b) To retain domain discrimination in the feature generation procedure, we encourage the real seen features to attract the generated seen features and repel the generated unseen features. (c) The combination of (a) and (b) constructs the final model. 
        } 
        \label{2}
\end{figure*}
\begin{figure*}[!ht]
\centering
  \includegraphics[width=.76\textwidth,height=8cm]{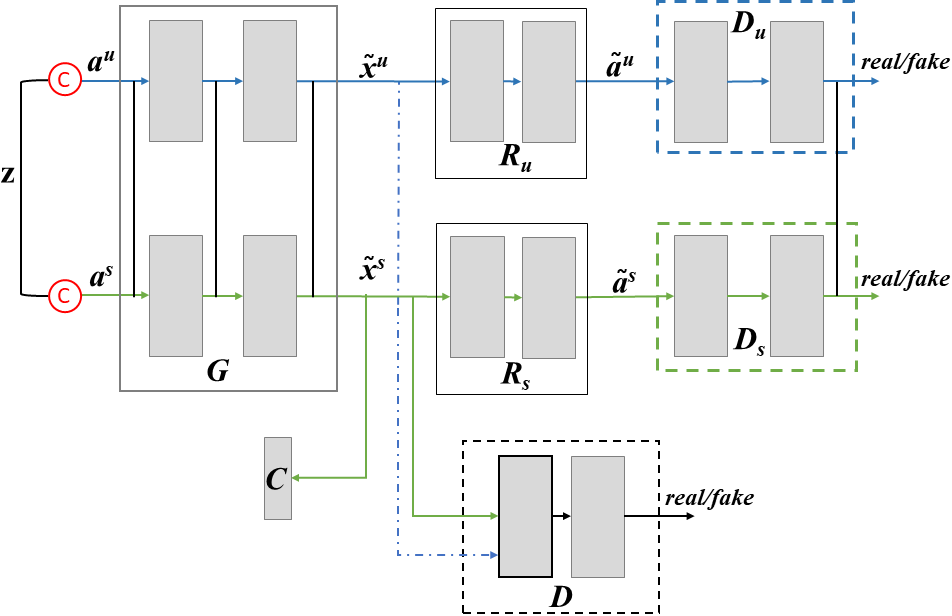}
\caption{Block diagram of our proposed BMCoGAN.}
\label{f1} 
\end{figure*}

\subsection{Proposed GZSL}
As mentioned earlier, the proposed BMCoGAN addresses two broad tasks: 1) Learning joint distribution through bidirectional mapping and 2) Learning to preserve discriminative information in synthesized features. 

Fig.~\ref{2}(a) shows an illustration of the concept for Task 1. For both seen and unseen domains, our aim is to learn the underlying semantic and visual distributions. We plan to construct visual features of both domains using the available class semantics. The quality of constructed seen class visual features can be evaluated by the available real features. However, for unseen classes, we do not have access to real visual features. Therefore, to evaluate, we utilize bidirectional mapping. This means we reconstruct the class semantics back from the generated features and then compare real semantics with reconstructed semantics. This bidirectional mapping not only helps to supervise unseen feature learning, also establishes a strong coupling between semantic and visual spaces. To fulfill our goal of learning the joint distribution, besides the unseen domain, we extend the bidirectional mapping to the seen domain.

For Task 2, compared to the constructed unseen features, we encourage the constructed seen features to remain closer to the real seen feature space. As shown in Fig.~\ref{2}(b), this objective is fulfilled by pushing the constructed seen features towards the real seen features and pulling the constructed unseen features away from the real seen features. The ultimate concept of the proposed model (Fig.~\ref{2}(c)) is a combination of both Tasks.

\textbf{Architecture Overview.} 
The proposed BMCoGAN is illustrated in Fig.~\ref{f1}. In the traditional CoGAN, the generators share only the bottom layers so that the GANs learn to deal with high-level concepts in the same manner. This also helps in learning the joint distribution of data samples. Unlike CoGAN, in BMCoGAN, we propose to use a fully shared generator for seen-unseen domains. So, the main component of BMCoGAN, the shared generator $G$ modifies the first constraint of CoGAN. We demonstrate that the fully shared generator is optimal than partially weight shared generators for GZSL tasks in the ablative section.

Moreover, contrary to CoGAN, our shared generator $G$ is conditioned on the class semantic vectors for generating visual features. The shared generator helps to learn to handle low to high-level features for both domains in a similar manner, which is crucial for our problem setting and encourages the joint distribution learning paradigm.
The generator $G$ plays an important role in accomplishing Tasks 1 and 2.  

To introduce bidirectional mapping within the CoGAN structure, we place two separate regressors ($R_s$ and $R_u$). The regressors map visual features to their corresponding class semantic vectors. Following the structure of CoGAN, we introduce partial weight shared coupled discriminators ($D_s$ and $D_u$) in the network. Contrary to CoGAN, the coupled discriminators' job is to learn to distinguish between the real and reconstructed class semantic vectors. We utilize adversarial loss to relate these components. In particular, the generator and the regressors combinedly generate class semantic vectors through bidirectional mapping and try to deceive the coupled discriminators. On the other hand, the coupled discriminators try to recognize real or generated class semantic vectors. 

The proposed method also has a feature discriminator $D$, which learns to separate real and generated visual features for the seen classes. The generator $G$ and discriminator $D$ interacts with each other adversarially. We adopt Wasserstein generative adversarial optimization for this supervision.

To learn domain discriminative information for handling bias towards seen classes, the generator $G$ and the discriminator $D$ learn to attract seen classes generated features to real seen classes features and repels unseen classes generated features away from real seen classes features. The classifier $C$ is placed to assess and enrich the discriminative characteristics in the generated features. The proposed model provides a unified framework to learn the joint distribution of seen-unseen classes simultaneously and retain domain distinctive information and can be trained in an end-to-end fashion.

\subsubsection{Learning joint distribution using bidirectional mapping}
We want to utilize the class semantic vectors of both seen and unseen classes and generate synthetic visual features. We follow \cite{xian2018feature} to construct the conditional Generator $G: \mathcal{Z}\times \mathcal{A}\rightarrow \mathcal{X}$ that uses random Gaussian noise $z\in \mathcal{Z}$ and class semantic vector $a_{c}\in \mathcal{A}$, and generates visual feature $x\in \mathcal{X}$ of class $c$. Now, the feature Generator $G$ can generate seen and unseen classes visual features conditioned on their class semantic vectors.

On the other hand, the regressors have to perform the reverse task of constructing the class semantic vectors from the generated visual features. We aim to train the regressors to generate class semantic vectors as similar as possible to the real class semantic vectors. This will ensure strong coupling between \textit{semantic}$\rightarrow$ \textit{visual} and \textit{visual}$\rightarrow$ \textit{semantic}. The semantic and visual spaces are better related through this bidirectional mapping. Note that we place separate regressors for the seen and unseen domains to encourage solid seen-unseen bidirectional mapping. We train the regressors with supervised loss as follows,
\begin{equation} \label{e2}
\begin{split}
\mathcal{L}_{Reg}=||a-\tilde{a}||_2^2.
\end{split}
\end{equation}
We compute $\mathcal{L}_{Reg}^s$ and $\mathcal{L}_{Reg}^t$ for $R_s$ and $R_u$, respectively. 

The ultimate aim of this task is to learn the joint distribution of seen and unseen domains. We optimize the shared generator $G$ to learn the joint distribution by adversarially interacting with coupled discriminators $D_s$ and $D_u$. The discriminator $D_s$ and $D_u$ receive reconstructed class semantics from the regressors $R_s$ and $R_u$, respectively and compares them with real semantics. The coupled discriminators have a weight-shared last layer. This reduces the number of parameters in the network. 
The adversarial optimization between the generator and the coupled discriminators is as follows,
\begin{equation} \label{e3}
\begin{split}
\min _{G} \max _{D_s, D_u} \mathcal{L}_{G1}&=\mathbb{E}_{{a^s}\sim p(a^s)}[\log D_s(a^s)]\\
&+\mathbb{E}_{{\tilde{a}^s}\sim p(\tilde{a}^s)}[\log (1-D_s(\tilde{a}^s))]
\\&+\mathbb{E}_{{a^u}\sim p(a^u)}[\log D_u(a^u)]\\
&+\mathbb{E}_{{\tilde{a}^u}\sim p(\tilde{a}^u}[\log (1-D_u(\tilde{a}^u))].
\end{split}
\end{equation}
Here, $\tilde{a}$ denotes the constructed class semantics by the regressors. This adversarial optimization encourages dual-domain joint distribution learning.

To impose the knowledge of the true underlying visual distribution in the generated visual space, we further adversarially supervise the generator with the real seen features. To be specific, we train the discriminator $D$ to distinguish between a real pair ($x^s,a^s$) and a synthetic pair ($\tilde{x}^s,a^s$) and the feature generator to produce synthetic features indistinguishable. The adversarial supervision is performed by using Wasserstein distance as follows \cite{xian2018feature},
\begin{equation} \label{e4}
\begin{split}
\min_{G} \max_{D} \mathcal{L}_{G_2}=& E_{p(x^s, a^s)}[D(x^s, a^s)]-E_{p(\tilde{x}^s, a^s)}[D(\tilde{x}^s, a^s)]-\\ & \beta E_{p(\hat{x}^s, a^s)}\left[\left(\left\|\nabla_{\hat{x}} D(\hat{x}^s, a^s)\right\|_{2}-1\right)^{2}\right],
\end{split}
\end{equation}
where $\hat{x}^s=\alpha x + (1-\alpha)\tilde{x}^s$ with $\alpha\sim U(0,1)$. $\beta$ is the penalty coefficient. Wasserstein distance is calculated using the first two terms. The gradient penalty is computed by the final term i.e., the gradient of the discriminator $D$ is forced to maintain a unit norm along the straight line between real and generated seen domain feature pairs.

\subsubsection{Learning to preserve discriminative information in synthesized features}
To preserve a distinction between the generated features of seen classes from that of the unseen classes, we explicitly design an optimization function, which encourages the generator $G$ to generate seen visual features closer to real seen visual features and generate unseen visual features away from real seen visual features. This phenomenon will help maintain a suitable gap between generated seen and unseen features for better classification and reduce the bias towards source classes. We optimize the following loss for holding discrimination,
\begin{equation} \label{e5}
\begin{split}
\mathcal{L}_{d}=||D(x^s)-D(\tilde{x}^s)||_2^2-||D(x^s)-D(\tilde{x}^u)||_2^2.
\end{split}
\end{equation}

We hypothesize that during testing, using the trained giscriminator $D$ will help to retain the domain distinctive information in the synthesized features. Thus, along with the Generator $G$, we extend the discriminative learning to the discriminator $D$. In particular, we optimize both $G$ and $D$ with the above loss with the hope to utilize the partial layers of $D$ in test time feature synthesis. 

To further support the discrimination among seen and unseen classes in the feature learning procedure, we constrain the early layers of the discriminator $D$ as follows,
\begin{equation} \label{e6}
\begin{split}
\mathcal{L}(D,\mathbb{C}) &=\mathbb{E}_{p(x^s, y^s)}\left[\mathbb{E}_{p_{D}(k \mid x^s)}\left[\mathcal{L}_{\mathbb{C}}\left(k, y, y^{\prime}\right)\right]\right]\\
\end{split}
\end{equation}\\
where, $\mathcal{L}_{ \mathbb{C}}\left(k, y, y^{\prime}\right)=\max \left(0, \Delta+\left\|k- \mathbb{C}_{y}\right\|_{2}^{2}-\left\|k- \mathbb{C}_{y^{\prime}}\right\|_{2}^{2}\right)$ \cite{wen2016discriminative}, $k$ is the early layer output of $D$ for the seen classes, $y$ and $y^{\prime}$ are the true class label of $x^s$ and random class label other than $y$, respectively. The $\mathbb{C}_{y}$ denotes the $y^{th}$ seen class center of deep features. The distributions of seen classes are grouped according to their labels by the center loss. This in turn facilitates the learned feature space to maintain distance among seen and unseen classes by concentrating the seen classes near their corresponding centers.

Finally, to make the generated features well suited for learning a discriminative target classifier, in line with \cite{xian2018feature}, we utilize the decision of a seen classes trained Classifier $C$ as,
\begin{equation} \label{e7}
\begin{split}
\mathcal{L}_{cls}&=-E_{\tilde{x}^s \sim p_{syn}}[\log P(y \mid \tilde{x}^s ; \theta)],
\end{split}
\end{equation}
where $y$ is the true class label of $\tilde{x}$ and $P(y \mid \tilde{x} ; \theta)$ denotes the probability of $\tilde{x}$ being predicted as $y$ conditioned on its semantic descriptor $a$. 

Our ultimate objective becomes,
\begin{equation} \label{e8}
\begin{split}
\min _{G, R_s, R_u} \max _{D_s,D_u} \lambda_{1}\mathcal{L}_{G1} + \mathcal{L}^s_{Reg} + \mathcal{L}^u_{Reg},\\ 
\min _{G} \max _{D}\lambda_{2}\mathcal{L}_{G2} + \lambda_{cls}\mathcal{L}_{cls} + \lambda_d\mathcal{L}_{d} + \lambda_{cen}\mathcal{L}(D,\mathbb{C}).
\end{split}
\end{equation}
Here, $\lambda_1$, $\lambda_2$, $\lambda_d$, $\lambda_{cls}$ and $\lambda_{cen}$ are hyper-parameters for weighting the losses. More detail on setting the hyper-parameters will be discussed in the results section. The training procedure of BMCoGAN is outlined in Algorithm \ref{algo}. 
\begin{algorithm}[!ht]
	\caption{Training Procedure}
	\label{algo}
\textbf{Input: }labeled seen dataset $\mathcal{Y}^s$; seen class semantic vectors $\mathcal{A}^s$; unseen class semantic vectors $\mathcal{A}^u$; generator $G$; regressors $R_s$ and $R_u$; coupled discriminators $D_s$ and $D_U$; discriminator $D$; seen classifier $C$ pre-trained on $\mathcal{Y}^s$. \\
\textbf{Output: }Trained generator $G$ and discriminator $D$. 
	\begin{algorithmic}[1]
	\While {not converged}
		\State Sample mini-batch from ${(x_i^s,y_i^s)}_{i=1}^{n_s}$ and their corresponding class semantic vectors $\mathcal{A}^s$;
		\State Sample mini-batch of class semantic vectors from $\mathcal{A}^u$;
		\State Compute losses (\ref{e2}) and (\ref{e3});
		\State Update $R_s$, $R_u$, $D_s$, and $D_u$ by (\ref{e2}) + (\ref{e3});
		\State Compute losses (\ref{e4}) -- (\ref{e6});
		\State Update $D$ by (\ref{e4}) + (\ref{e5}) + (\ref{e6});
		\State Compute losses (\ref{e3}), (\ref{e4}), (\ref{e5}), and (\ref{e7}) for $G$;
		\State Update $G$ by (\ref{e3}) + (\ref{e4}) + (\ref{e5}) + (\ref{e7});
		\EndWhile\\
	\Return {Trained $G$ and $D$.} 
	\end{algorithmic} 
\end{algorithm}
\begin{figure}[!ht]
\centering
  \includegraphics[width=\columnwidth,height=2.8cm]{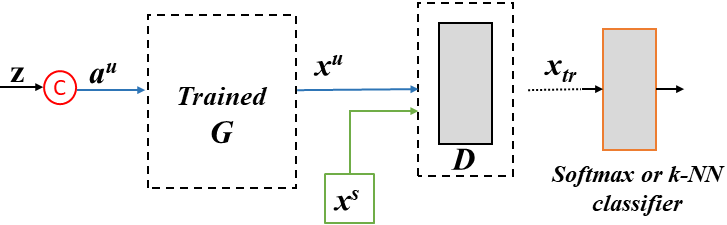}
\caption{Testing sequence of our proposed GZSL method.}
\label{test} 
\end{figure}

\textbf{Testing Phase :} The test phase of the proposed method is shown in Fig.~\ref{test}. We generate multiple synthetic features for every unseen class by utilizing the trained generator $G$ and re-sampling the noise $z$. Then, to retain the learned distinctive characteristics in the final test time training dataset's feature space, we pass the seen domain test features and generated unseen domain features through the early layers of trained $D$. Once we have training data for both seen $x^s_{tr}$ and unseen classes $x^t_{tr}$, we learn a classifier separately as the final supervised classifier. In this work, we evaluate two kinds of classifiers: softmax and k-nearest neighbor (k-NN).
\begin{table*}[!ht]
\caption{Performance comparison. U and S are the Top-1 accuracies tested on target classes and source classes, respectively, in GZSL. H is the harmonic mean of U and S. $\triangle$, $\square$, and $\square \square$ denote embedding learning, generative, and bidirectional mapping methods, respectively, and `-' represents that the results were not reported.}
	\begin{center}
	\resizebox{.9\textwidth}{!}{%
		\begin{tabular}{c|c|cccccccccccc}
			\hline
			\multirow{3}{*}{Approach} &\multirow{3}{*}{Model}&  \multicolumn{12}{c}{GZSL}\\\cline{3-14}
		 && \multicolumn{3}{c}{CUB} & \multicolumn{3}{c}{SUN} & \multicolumn{3}{c}{AWA1} &\multicolumn{3}{c}{AWA2} \\ 
			&&$U$& $S$& $H$& $U$& $S$& $H$& $U$& $S$& $H$& $U$& $S$& $H$\\\hline\hline
			\multirow{7}{*}{$\triangle$} 
			&DCN \cite{liu2018generalized} (2018)
			&28.4&60.7&38.7&25.5&37.0&30.2&25.5&84.2&39.1&-&-&-\\
			&CRnet \cite{zhang2019co} (2019)
			& 45.5 &56.8& 50.5 &34.1& 36.5& 35.3&58.1& 74.7& 65.4&-&-&-\\
			&TCN \cite{jiang2019transferable} (2019)
			&52.6&52.0&52.3&31.2&37.3&34.0&49.4&76.5&60.0&61.2&65.8&63.4\\
			&DVBE \cite{min2020domain} (2020)
			&53.2& 60.2 &56.5&45.0& 37.2& 40.7&-&-&-&63.6 &70.8& 67.0\\
			&DAZLE \cite{huynh2020fine} (2020)
			&56.7&59.6&58.1& 52.3&24.3&33.2&-&-&-&60.3&75.7&67.1\\
		    &VSG-CNN \cite{geng2020guided} (2020)
		    & 52.6& 62.1& 57.0&30.3 &31.6& 30.9&-&-&-&60.4& 75.1& 67.0\\
		    \hline
			\multirow{10}{*}{$\square$}&SE-GZSL \cite{kumar2018generalized} (2018)
			& 41.5& 53.3& 46.7&40.9& 30.5& 34.9&56.3& 67.8 &61.5&58.3& 68.1& 62.8\\
			&f-CLSWGAN \cite{xian2018feature} (2018)
			&43.7 &57.7& 49.7&42.6 &36.6& 39.4&57.9 &61.4& 59.6&-&-&-\\
			&cycle-CLSWGAN \cite{felix2018multi} (2018)
			& 45.7& 61.0& 52.3&49.4& 33.6& 40.0&56.9& 64.0& 60.2&-&-&-\\
			&CADA-VAE \cite{schonfeld2019generalized} (2019)
			& 51.6& 53.5& 52.4&47.2 &35.7& 40.6&57.3& 72.8& 64.1&55.8&75.0&63.9\\
			&f-VAEGAN-D2 \cite{xian2019f} (2019)
			& 48.4 &60.1 &53.6&45.1 &38.0& 41.3&-&-&-&57.6&70.6&63.5\\  
			&LisGAN \cite{li2019leveraging} (2019)
			&46.5 &57.9 &51.6&42.9& 37.8 &40.2&52.6& 76.3& 62.3&-&-&-\\
			&GMN \cite{sariyildiz2019gradient} (2019)
			&56.1 &54.3 &55.2&53.2& 33.0& 40.7&61.1& 71.3& 65.8&-&-&-\\
			
			&RZSL-CVCP\cite{li2019rethinking} (2019)			&47.4&47.6&47.5&36.6&42.8&39.3&62.7&77.0&69.1&56.4&81.4&66.7\\
			&RFF-GZSL (softmax) \cite{han2020learning} (2020)
			& 52.6 &56.6 &54.6&45.7& 38.6 &41.9&59.8& 75.1& 66.5&-&-&-\\
			&LsrGAN \cite{vyas2020leveraging} (2020)&48.1&59.1&53.0&44.8&37.7&40.9&54.6&74.6&63.0&-&-&- \\
			&TF-VAEGAN \cite{narayan2020latent} (2020)&52.8&64.7&58.1&45.6&40.7&43.0&59.8&75.1&66.6&-&-&- \\
			&TI-GZSL(Res) \cite{feng2020transfer} (2020)&44.8&42.2&43.5&31.5&20.3&24.7&61.5&67.7&64.4&72.1&63.9&67.7\\
			&ASPN \cite{lu2020attentive} (2020)&50.7&61.5&55.6&-&-&-&58.0&85.7&69.2&46.2&87.0&60.4\\
			
		\hline
			\multirow{10}{*}{$\square \square$}&DASCN \cite{ni2019dual} (2019)
			&45.9&59.0&51.6&42.4&38.5&40.3&59.3&68.0&63.4&-&-&-\\
			&GDAN \cite{huang2019generative} (2019)
			&39.3&66.7&49.5&38.1&89.9&53.4&-&-&-&32.1&67.5&43.5\\
			&RBGN \cite{xing2020robust} (2020)&47.0&54.3&50.4&46.0&37.2&41.2&57.5&67.1&61.9&57.2&71.4&63.6\\
			&GZSL-AVSI \cite{chandhok2020enhancing} (2020)	&61.2&57.7&59.4&-&-&-&60.5&71.9&65.7&59.4&74.2&66.0\\
			&ISE-GAN \cite{pambala2020generative} (2020)	&52.4&55.4&53.8&51.3&34.7&41.4&58.7&74.4&65.6&55.9&79.3&65.5\\
			&SE-GAN+SM \cite{pambala2020generative} (2020)&48.4&57.6&52.6&44.7&37.0&40.5&53.9&68.3&60.3&55.1&61.9&58.3\\
			&IBZSL \cite{liu2020information} (2020)&52.2 &56.2 &54.1&43.8& 37.8& 40.6&-&-&-&56.0& 80.0 &65.9\\
			&cFlow-ZSL \cite{gu2020generalized} (2020) &50.8 &54.9 &52.8&46.7 &39.5& 42.8&57.1& 68.1& 62.1&56.7 &74.8& 64.5\\
			\cline{2-14}
			&BMCoGAN (softmax) &57.9&66.1&61.7 &52.9&43.7&47.8 &61.5&78.2&68.8 &61.9&76.9&68.5\\
			&BMCoGAN (1-NN) &64.6&73.5&\textbf{68.7} & 58.1&52.4&\textbf{55.1} &66.1&86.1&\textbf{74.7} &66.9&81.3&\textbf{73.4}\\
		\hline
		
		\end{tabular}
		}
	\end{center}
	
	\label{t1}
\end{table*}

For softmax classifier, after receiving features from early layers of $D$, the classifier produces $|C^s + C^t|$ dimensional output, i.e., the $|C^s|$ seen classes and $|C^t|$ unseen classes. During the forward pass, within $h_2$, the features are transformed to a $|C^s + C^t|$-dimensional class probability through log softmax function.
We define the classification loss as follows,
\begin{equation} \label{e9}
\begin{split}
\mathcal{L}_{h}&=-E_{ x_{tr} \sim p_{tr}}[\log P(y \mid x_{tr} ; \theta_{h_2})]
\end{split}
\end{equation}
where, $x_{tr}$, $y$, and $p_{tr}$ denote the samples of the newly formed training dataset, the true class label of $x_{tr}$, and distribution of the new training dataset respectively. $P(y \mid x_{tr} ; \theta_{h_2})$ represents the probability of $x_{tr}$ being recognized as $y$. 

For the k-NN classifier, we follow a similar setting as \cite{han2020learning}. We only evaluate using 1-NN classifier.

\section{Experimental Studies}
In this section, we describe the datasets, evaluation metrics, implementation details, discuss our experimental outcomes (classification results, hyper-parameters setups, and number of synthesized features per unseen class) and present a detailed ablation study.

\textbf{Datasets:} We conduct our experiments on four popular datasets, Caltech-UCSD Birds-200-2011 (CUB) \cite{welinder2010caltech}, SUN Attribute (SUN) \cite{patterson2012sun}, Animals with Attributes 1 (AWA1) \cite{lampert2009learning}, and Animals with Attributes 2 (AWA2) \cite{xian2018zero}. We follow \cite{xian2017zero}, to split the total classes into source and target classes on each dataset.

\textbf{CUB} contains a total of 11,788 images of 200 classes of fine-grained bird species. The dataset has 312 annotated attributes for every class. Among the classes 150 are selected as source classes and the remaining 50 classes are treated as target or unseen classes. \textbf{SUN} is composed of 14,340 images of 717 different categories of scenes. The number of source and target classes used for GZSL is 645 and 72 respectively. The dataset has 102 annotated attributes and only 16 images per class. \textbf{AWA1} consists of 30,475 images of animals of 50 different classes with 85 annotated attributes. For GZSL, 40 classes are used as source and 10 are used as target classes. \textbf{AWA2} has 40 source and 10 target classes comprising 37,322 images in total.

\textbf{Evaluation Metrics: }We evaluate the performance of our method by per-class Top-1 accuracy. For the source domain, we will evaluate the Top-1 accuracy on source classes denoted as $S$. For the target domain, the Top-1 accuracy on the target classes is represented as $U$. For evaluating the total performance of GZSL, we compute the harmonic mean in line with \cite{xian2017zero} as,
\begin{equation} \label{e10}
\begin{split}
H&=\frac{2\times S\times U}{S + U}
\end{split}
\end{equation}

\textbf{Implementation Details: }In our experiments, we extract $2048$ dimensional features from pre-trained ResNet-101 for all seen classes. Since the generator has to produce fully-connected features from conditional input, we maintain a total fully-connected structure of the generator for efficiency i.e., the generator has one hidden fully-connected layer of dimension $4096$. The fully-connected structure of the generator also helps in learning the joint distribution of the seen-unseen domains. Both the Regressors have a hidden layer with $1024$ fully-connected neurons. The coupled discriminators reduce the class semantic vectors to $256$ dimension through separated fully-connected layers and then share the final fully-connected layer. 

We observed that the discriminator $D$ co-operates more in the discriminative learning and supervision when it has only one hidden layer. We have used $1024$-dimensional fully-connected hidden layer for the discriminator $D$ in our experiments. We follow \cite{gulrajani2017improved} for improved Wasserstein GAN training. Adam solver with $\beta_1=0.5$ and $\beta_2=0.999$ is used for optimization. A learning rate of $0.0001$ is used for the generator, the discriminator $D$, and the center loss (\ref{e6}), and $0.0002$ is used for the regressors and the coupled discriminators. For our final objective, we find setting $\lambda_1=2$, $\lambda_2=0.8$, $\lambda_d=1$, $\lambda_{cls}=0.2$, and $\lambda_{cen}=0.1$ optimal.

\subsection{Results and Analysis} 
We compare BMCoGAN with state-of-the-art methods on generalized zero-shot learning, and the results are shown in Table~\ref{t1}. The results of LATEM \cite{xian2016latent}, DEM \cite{zheng2017learning}, and SGMAL \cite{zhu2019semantic} are adopted from SGMAL \cite{zhu2019semantic} and the results of other compared methods are obtained from their corresponding published articles. We compare the performance with only inductive methods for a fair comparison, which do not use unseen images during training. The results with $400$ synthesized features per class are shown in Table~\ref{t1}. Though we present results from embedding learning and feature synthesizing methods in Table~\ref{t1}, for comparison, our main focus is on the bidirectional methods. We perform GZSL tasks with the proposed BMCoGAN using both $1$-NN and softmax classifier. The Harmonic mean (\ref{e10}) is the main indicator of how well a GZSL method performs. 

Table~\ref{t1} shows that BMCoGAN with softmax classifier outperforms all the contemporary bidirectional mapping GZSL methods significantly in terms of the Harmonic mean $H$. This variant also achieves better $U$ accuracy than the bidirectional and contemporary embedding learning and feature synthesizing methods for most tasks. BMCoGAN with 1-NN classifier achieves better $H$ accuracy than all compared bidirectional and other methods for all tasks. This variant also outperforms the majority of contemporary methods in both $U$ and $S$ accuracy for all four datasets. We observe that the 1-NN classifier variant of BMCoGAN performs better than the softmax classifier variant, which means the learned feature space is more suitable for nearest neighbour classifier.

Note that, in general, BMCoGAN maintains a good balance between the $S$ and $U$ accuracies. That is, besides achieving higher $U$ accuracy than other bidirectional mapping methods, it does not show a significant drop in $S$ accuracy such as \cite{chandhok2020enhancing} (CUB). This means the proposed method can generate well discriminative features for improved dual-domain classification. Both versions of BMCoGAN achieve improved performance than ISE-GAN \cite{pambala2020generative} and IBZSL \cite{liu2020information}, and verify better discriminative learning, which reduces bias towards seen classes.

\textbf{Analyzing Number of Generated Features} \begin{figure}[!ht]
    \centering
    \subfigure[CUB]{\includegraphics[width=0.4\columnwidth]{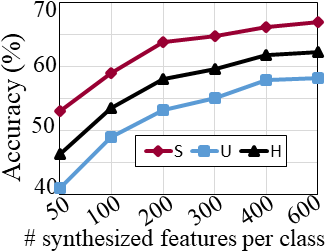}}
    \subfigure[SUN]{\includegraphics[width=0.4\columnwidth]{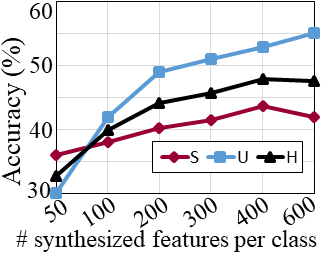}}
    \\
     \subfigure[CUB]{\includegraphics[width=0.4\columnwidth]{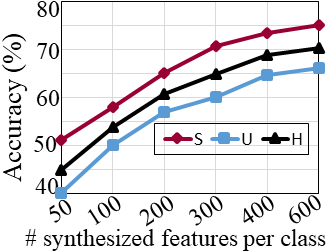}}
    \subfigure[SUN]{\includegraphics[width=0.4\columnwidth]{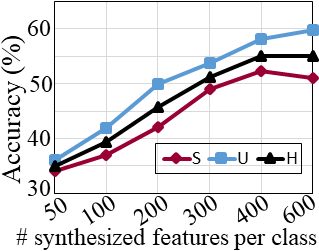}}
    \caption{(a) and (b) Increasing the number of synthesized features wrt GZSL performance in CUB and SUN datasets using \textbf{softmax classifier}. (c) and (d) Increasing the number of synthesized features wrt ZSL performance in CUB and SUN datasets using \textbf{1-NN classifier}.}
    \label{gzslsamples}
\end{figure}
For analyzing the effect of the number of generated features per class during testing, we plot the graphs in Fig~\ref{gzslsamples}. The graphs show the performance comparison of CUB and SUN datasets for various generated features per class. In general, we demonstrate that with the increasing number of features per class, the harmonic mean $H$ of both datasets increases. 

For the softmax version, in the CUB dataset, $S$ and $U$ significantly increase till $400$, and after that, the increment is marginal (Fig~\ref{gzslsamples}(a)). For the SUN dataset, $S$ decreases after $400$; however, $U$ increases with the increasing number of features per class (Fig~\ref{gzslsamples}(b)). Notice that after $400$, the value of harmonic mean plateaus as both $S$ and $U$ show no significant changes.

To demonstrate the trend of GZSL performance of 1-NN variant of BMCoGAN on a different number of generated features per class, we plot the graph shown in Figs~\ref{gzslsamples}(c) and \ref{gzslsamples}(d). The CUB dataset shows consistently increasing performance wrt an increasing number of synthesized features per class. On the other hand, the SUN dataset shows plateaued $H$ accuracy with slightly increasing $S$ and decreasing $U$ accuracies. Overall, we observe that synthesizing $400$ to $600$ features per class is suitable for BMCoGAN, depending on the dataset.

\begin{figure}[!ht]
    \centering
    \subfigure[]{\includegraphics[width=0.4\columnwidth]{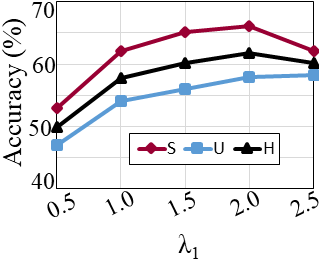}} 
    \subfigure[]{\includegraphics[width=0.4\columnwidth]{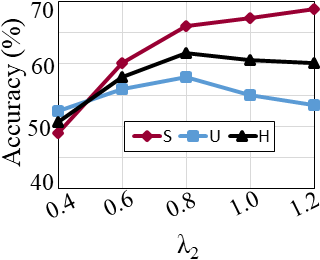}}\\ \subfigure[]{\includegraphics[width=0.4\columnwidth]{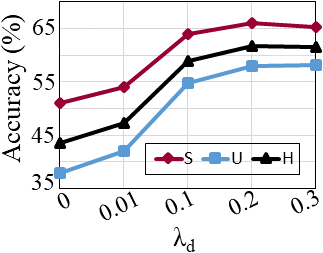}}
    \subfigure[]{\includegraphics[width=0.4\columnwidth]{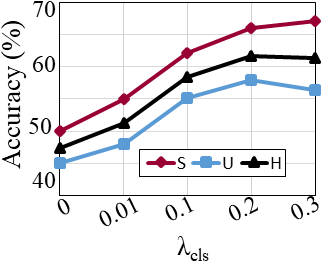}}\\
    \subfigure[]{\includegraphics[width=0.4\columnwidth]{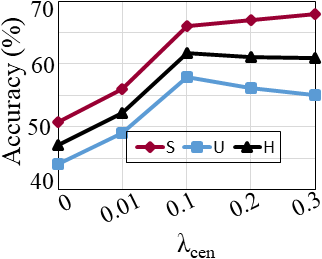}}
    \caption{Effect of varying the hyper-parameters in the GZSL performance on the CUB dataset.}
    \label{parameters}
\end{figure}
\begin{table*}[!ht]
\caption{Ablative analysis for GZSL on the CUB dataset.}
	\begin{center}
	\resizebox{.47\textwidth}{!}{%
		\begin{tabular}{c|ccc}
			\hline
			&\multicolumn{3}{c}{CUB} 
			\\\cline{2-4}
		 	\multirow{1}{*}{Approach}& U& S& H
			\\\hline\hline
			BMCoGAN w/o $\mathcal{L}_{G2}$ &48.3 &44.2&46.1\\
			BMCoGAN w/o $\mathcal{L}_{d}$&38.7&51.9&44.3\\
			BMCoGAN w/o $\mathcal{L}_{cls}$&45.8&50.7&48.1\\
			BMCoGAN w/o $\mathcal{L}_{cen}$&44.8&50.8&47.6\\
			BMCoGAN w/ $R$ &53.4 &64.1&58.2\\
			BMCoGAN w/ sep. $D_s$ and $D_u$ &56.1 &65.2&60.3\\
			BMCoGAN w/ sep. $G_s$ and $G_u$&45.1&54.9&49.5\\
			BMCoGAN w/o $D$ (test) &52.1&65.2&57.9\\
			BMCoGAN &\textbf{57.9}&\textbf{66.1} &\textbf{61.7}
			\\\hline
		\end{tabular}
		}
	\end{center}
	
	\label{t3}
\end{table*}

\textbf{Hyper-parameters Analysis.} For studying the trend of GZSL accuracy of BMCoGAN in different hyper-parameters ($\lambda_1$, $\lambda_2$, $\lambda_{cls}$, and $\lambda_{cen}$) settings, we plot the graphs shown in Fig~\ref{parameters} for CUB dataset. We present the analysis of the hyper-parameters setting for the softmax classifier variant of the proposed method.

The hyper-parameter $\lambda_1$ weights the contribution of bidirectional adversarial learning in the overall optimization and joint distribution learning. Fig~\ref{parameters}(a) shows the performance of BMCoGAN with various $\lambda_1$ setups. We observe that seen accuracy increases with the increase of $\lambda_1$ till $\lambda_1=2$ and decreases after that. On the other hand, the increment rate of unseen accuracy plateaus after $\lambda_1=2$. However, it does not show a decreasing pattern. Though the gap between seen and unseen accuracy decreases after $\lambda_1=2$, the $H$ accuracy decreases. Thus, we find the value of $\lambda_1=0.2$ optimal as the $H$ accuracy is optimal at that value. Overall, the graph indicates that a significant contribution of $\mathcal{L}_{G1}$ reflects better GZSL recognition and verifies the robustness of the proposed BMCoGAN network for learning better dual-domain distribution. 

The supervisory WGAN optimization is weighted by $\lambda_2$. As shown in Fig~\ref{parameters}(b), the seen accuracy shows an increasing pattern with higher $\lambda_2$ values. However, the unseen accuracy depicts a sharp decrement after $\lambda_2=0.8$. It is worth noting that the unseen accuracy slightly surpasses the seen accuracy at a significantly low $\lambda_2$ value. However, with the increasing contribution of $\mathcal{L}_{G2}$, the network becomes more biased towards seen classes. The network shows optimal $H$ accuracy at $\lambda_2=0.8$. The optimum supervision of $\mathcal{L}_{G2}$ is important to maintain a proper balance between $S$ and $U$ accuracies in the network.

To study the effect of discriminative loss $\mathcal{L}_d$, we plot the graph shown in Fig~\ref{parameters}(c). We observe that the unseen accuracy is significantly low, and the gap between seen and unseen accuracy is huge at near-zero $\lambda_d$ value. This means the source domain is dominating the classification performance. However, the high bias towards source classes reduces with the increasing value of $\lambda_d$, and the gap between seen and unseen accuracies also reduces. This evaluates the network is capable of preserving domain distinctive information with an optimal contribution of $\mathcal{L}_d$.

Figs~\ref{parameters}(d) and \ref{parameters}(e) show a similar trend. The seen accuracy depicts an increasing pattern, while the unseen accuracy decreases after a certain point. We demonstrate that up to a certain value, $\lambda_{cls}$ and $\lambda_{cen}$ serve their purpose of supporting the domain discrimination by preserving distinctive seen domain information in the network. Therefore, we set the value of $\lambda_{cls}$ and $\lambda_{cen}$ to $0.2$ and $0.1$, respectively for the optimal $H$ accuracy.

\textbf{Ablative Analysis.} To justify the role of different components and optimization in BMCoGAN separately, we present the ablative analysis in Table~\ref{t3}. We perform the ablative analysis by deducting vital components from BMCoGAN and introducing alternative components in BMCoGAN, and justify their impact on the performances.

The variant BMCoGAN w/o $\mathcal{L}_{G2}$ represents optimizing the BMCoGAN without the WGAN loss or supervision for the seen visual features. We observe that both seen and unseen accuracies significantly decrease, which means the supervision from real visual features is crucial for learning to generate visual features. Compared with unseen accuracy, the seen accuracy drastically decreases. This indicates the network loses the capacity to learn the true underlying distribution of the seen domain without supervision.

BMCoGAN w/o $\mathcal{L}_{d}$ denotes the variant without the optimization for retaining domain discriminative information in the synthesized features. The performance of this variant shows that without $\mathcal{L}_{d}$ optimization, the learned visual distribution is dominated by the seen classes. This creates towards seen classes. This justifies the importance of $\mathcal{L}_{d}$ in the network.

We demonstrate that the variants without $\mathcal{L}_{cls}$ and $\mathcal{L}_{cen}$ also show a significant drop in performance especially, for seen classes. This proves that these losses help the network learn distinctive knowledge about the seen classes, confine the generated seen features to their corresponding centers, and eventually reduce bias towards seen classes.

BMCoGAN w/ $R$ is the variant in which we replace separate regressors with a shared regressor $R$. The reduced performance indicates that learning separate bidirectional mapping for seen-unseen domains enhances joint distribution learning and GZSL recognition. 
The variant with separate class semantic discriminators is denoted as BMCoGAN w/ sep. $D_s$ and $D_u$. The slight decrement in the performance indicates that separate discriminators do not harm the network to a great extent, like other variants. However, the coupled discriminators encourage improved performance and reduce the number of parameters in the network.

BMCoGAN w/ sep. $G_s$ and $G_u$, this variant has coupled generators for seen-unseen domains similar to CoGAN \cite{liu2016coupled}. We observe that the performance drastically decreases. This means, only learning the underlying high-level concept in a similar way for both domains is not sufficient for the GZSL recognition task. Thus, the proposed structure of the shared generator is optimal for the task.

To verify the contribution of partial layers of $D$ in retaining the domain discrimination, we test the network without passing generated unseen features through early layers of $D$ in the variant BMCoGAN w/o $D$. We notice that not only the performance of GZSL decrease, the gap between seen and unseen accuracy significantly increases. This proves increased bias towards seen classes and less domain discrimination in the generated visual features.

\section{Conclusion}
In this work, we propose a new bidirectional mapping generative model for the generalized zero-shot learning tasks. In particular, we propose to incorporate the concept of bidirectional mapping into the coupled generative adversarial network for learning dual-domain (seen-unseen) joint distribution and design a loss optimization for preserving domain discrimination. We present and discuss the evaluation of our proposed methods on different benchmark datasets and the performance comparison with existing generalized zero-shot learning methods. We demonstrate that the proposed method outperforms contemporary methods. In addition, we provide detailed ablative analysis to discuss the importance of different components in the proposed network.


%





\ifCLASSOPTIONcaptionsoff
  \newpage
\fi

\bibliographystyle{IEEEtran}
\bibliography{ref}




\end{document}